\documentclass[runningheads]{llncs}
\usepackage[T1]{fontenc}
\usepackage{wrapfig}
\usepackage[most]{tcolorbox}
\usepackage{xcolor} 
\usepackage{stmaryrd}
\usepackage{amsmath,amsfonts,amsbsy}
\usepackage{xcolor}
\usepackage[mathscr]{euscript}
\usepackage{dutchcal}
\usepackage[ruled,linesnumbered,vlined,shortend]{algorithm2e}
\usepackage{algorithmicx}
\usepackage{graphicx}
\usepackage{caption}
\usepackage{subfigure}
\usepackage{booktabs}
\usepackage{multirow}
\usepackage{tabularx}
\usepackage{tablefootnote}
\usepackage{threeparttable}
\usepackage{mdwlist}
\usepackage{makecell}
\usepackage{diagbox}
\usepackage{relsize}
\usepackage{makecell}
 \usepackage[disable]{todonotes}
\usepackage[T1]{fontenc}
\usepackage{lipsum}
\usepackage{hyperref}
\usepackage{url}
\usepackage{enumitem}
\usepackage[misc]{ifsym}

\usepackage[capitalise,nameinlink]{cleveref}
\crefname{section}{Sec.}{Sec.}
\Crefname{section}{Section}{Sections}
\crefname{listing}{List.}{List.}
\crefname{listing}{Listing}{Listings}
\Crefname{listing}{Listing}{Listings}
\crefname{lstlisting}{Listing}{Listings}
\Crefname{lstlisting}{Listing}{Listings}

\newcommand{\hide}[1]{}

\newcommand{\fullmethod}[0]{\textsc{Actionable and Explainable Tensor Analysis System}\xspace}
\newcommand{\method}[0]{\textsc{AnTenA}\xspace}

\begin{document}
\title{AnTenA: Actionable and Explainable Tensor Analysis System with Large Language Models}
%
%

\author{Dawon Ahn\inst{1}\orcidID{0000-0002-4735-0337} \and
Audrey Der\inst{2}\orcidID{0000-0002-9266-5864} \and
Evangelos E. Papalexakis\inst{1}\orcidID{0000-0002-3411-8483}\Letter}
\authorrunning{Ahn et al.}
%
\institute{University of California, Riverside CA 92507, USA \\
\email{dahn017@ucr.edu, epapalex@cs.ucr.edu}\\
\and
Neuralix AI, Houston, TX, USA\\
\email{audrey@neuralix.ai}
}

\toctitle{AnTenA: Actionable and Explainable Tensor Analysis System with Large Language Models}
\tocauthor{Dawon Ahn, Audrey Der, Evangelos E. Papalexakis}
\maketitle              

\begin{abstract}
Accurately explaining hidden patterns in multi-aspect data has typically been done by leveraging labels and/or accompanying auxiliary metadata.
However, labels and auxiliary data may be inaccurate (e.g. nonstandard, inconsistent), insufficient (e.g. static tabular metadata for time-dependent recordings), or unavailable. 
%
We propose \fullmethod (\method), which leverages the knowledge of large language models (LLMs) to explain the hidden patterns in human narratives. 
\method uses task-agnostic and task-specific prompts to explain extracted co-clustered latent patterns from tensor decomposition.
To evaluate these explanations, we test the LLMs on forward and backward inference tasks. 
Our demo system is available at \url{https://github.com/dawonahn/ECML_PKDD_AnTenA}.

\keywords{Tensor Analysis  \and Explainability \and Large Language Model.}
\end{abstract}

\section{Introduction, Motivation, and Related Work}		\label{sec:intro}
Explainable AI (XAI) aims to explain the decisions of artificial intelligence (AI) models, but a large proportion of prior works use varying explanation modes, are supervised frameworks, and attempt to explain a black-box structure. Explanations for supervised models naturally explain \textit{how} models come to a decision within a black-box structure, not necessarily insight into the \textit{data}. Our problem space is in tensor data mining, and examples of these tensor XAI works include vision-based~\cite{yanglanguage}, graph-based~\cite{pan2024tagexplainer}, and feature-based~\cite{lengerich2023llms} explanations. 

\textbf{Low-Rank Latent Components.}
A tensor is a natural way to represent multi-faceted interactions, and is often \textit{unsupervised}. 
Tensor decomposition is a fundamental model to analyze tensors, which decomposes them into low-rank latent components. 
These components contain \textit{co-clustered} patterns across modes. 
By examining these latent components instead of the entire tensor, 
	people can discover hidden patterns across a wide range of domains~\cite{sidiropoulos2017tensor}.

%
\textbf{Explaining Latent Components.}
Prior works~\cite{park2016bigtensor,ahn2020gtensor} have used metadata to explain the latent components. However, like with  labels, we often cannot assume real-world datasets' auxiliary metadata is available or comprehensively captures the semantics of latent components. Solely relying on auxiliary information can be limited or misaligned with underlying dynamics (e.g. stale metadata).

%
\textbf{LLM-based Explainability.}

Recent work has shown that natural language explanations are more intuitive and directly interpretable compared to vision and graph-based alternatives, and are quickly becoming the preferred explanation format~\cite{wu2024usable}.
Latent components are less directly interpretable
because they capture multi-way interactions.
Studies have shown leveraging the reasoning capabilities and knowledge of LLMs as explainers is viable, because they are trained on vast amounts of information and can identify nuances~\cite{bilal2025llms}.
Our method proposes using an LLM to explain these components in natural language.


\textbf{Contribution.} We present an LLM-based explainable tensor analysis demo system that explains extracted latent components. We enhance the system with an interpretable neural tensor model for accuracy. Motivated by this work~\cite{yang2025llm}, we introduce forward and backward inference methods to improve the reliability of LLM explanations.

\vspace{-3mm}
\vspace{-3mm}
\section{System Overview}	\label{sec:method}
\vspace{-3mm}
\method{} consists of two modules: a tensor analysis and an explainer.

\textbf{Module 1: Tensor Analysis.}
This module extracts interpretable latent components using tensor decomposition models.
These models decompose a tensor into a set of latent components that capture hidden patterns between modes.
CANDECOMP/PARAFAC (CP)~\cite{kolda2009tensor} is widely used due to its simplicity and interpretability, but is limited to multi-linear operations.
To address this limitation, NeAT~\cite{ahn2024neural} was proposed to maintain the CP structure, while capturing non-linear latent patterns by applying neural networks to each latent component.
To improve the interpretability, we introduce non-negativity and orthogonality constraints into the models. The non-negativity constraint ensures all factor values as non-negative, making factor value magnitudes meaningful, and the orthogonality one ensures factors are distinct. 

\textbf{Module 2: Explainer.}
This module uses LLMs to explain the extracted components.
The inputs to LLM include top-$K$ entities for each mode and their corresponding factor values, decomposition results such as reconstruction error and factor match score (FMS)~\cite{acar2011all}, and optional meta-data. 
We have task-agnostic and task-specific prompts. 
Task-agnostic prompts instruct LLM to describe the hidden patterns, which allows the explanation to remain general. However, these explanations can be less actionable for specific tasks. In contrast, task-specific prompts guide the LLM to describe the patterns regarding given tasks. For example, when using MovieLens dataset in our demo system, we guide the LLM to suggest movie recommendations based on each component. 
These explanations can provide more actionable insights, but they might be biased to the given tasks. 
To evaluate the reliability of generations, we introduce two inference tasks, motivated by this work~\cite{yang2025llm}.
In the forward task, LLM is asked to predict entities that should belong to the given component but do not appear among its top-$K$ components. 
Providing the full entity list can lead to long prompts, so we construct an entity pool using top-50 entities across all components.
In the backward task, LLM is asked to predict an injected false entity within a component.
False entities are chosen from the top-$K$ entities of a distinct component, where its distinctiveness is measured with FMS score.
\

\vspace{-10mm}
\section{Demo}		\label{sec:experiments}
\begin{figure*} 
\centering
	\includegraphics[width=0.8\linewidth]{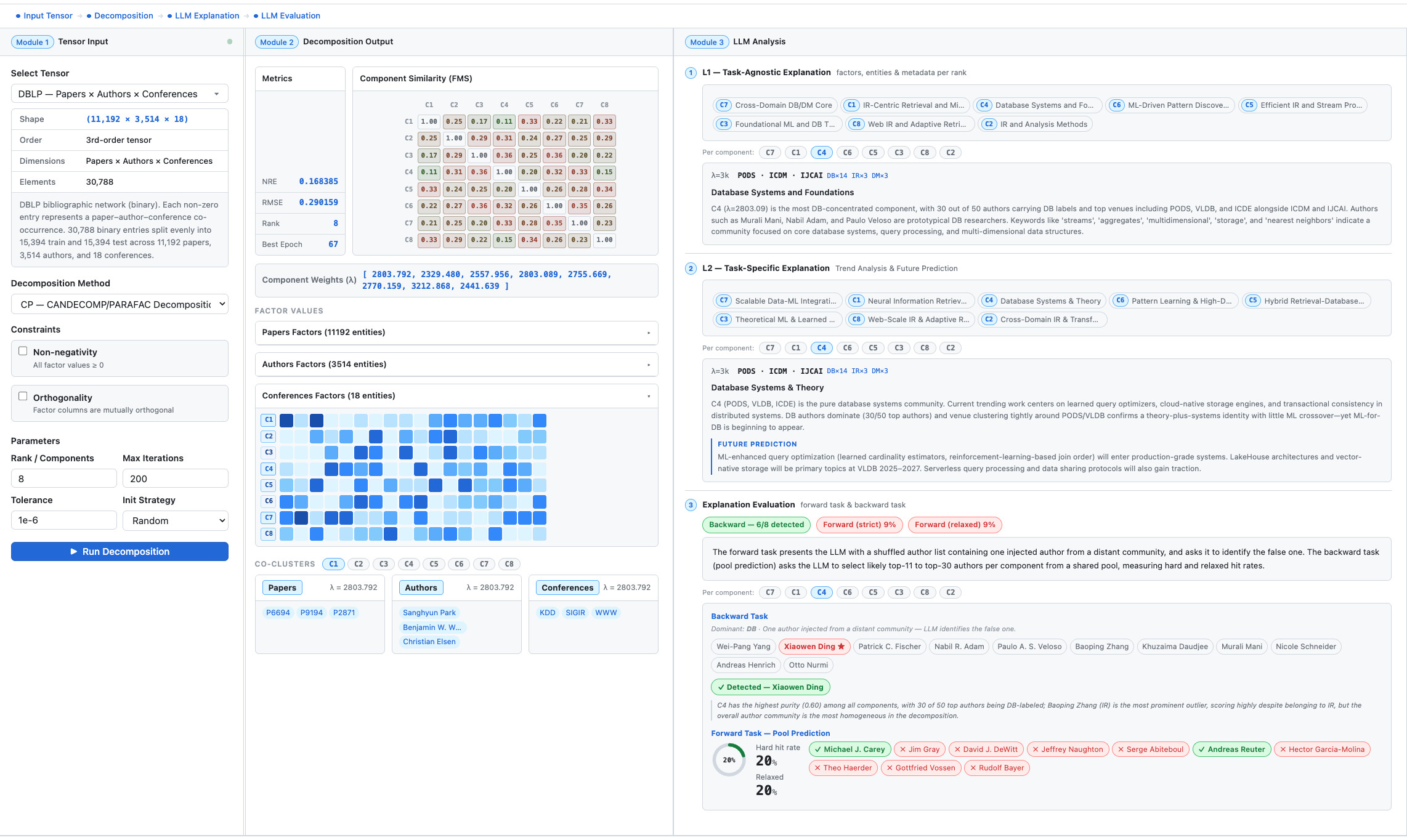}
	\caption{\label{fig:demo} Overview of \method using DBLP dataset and a CP model.}
\end{figure*}
We explain our demo system using a DBLP dataset as illustrated in \Cref{fig:demo}.
The first column allows the user to choose tensor datasets and models.
We showcase with two real-world tensors, MovieLens-Belief-2024\footnote{\url{https://grouplens.org/datasets/movielens/ml_belief_2024}} and DBLP\footnote{\url{https://github.com/Jhy1993/HAN/tree/master/data/DBLP_four_area}}.
The users can choose two tensor models, CP and NeAT, with constraint options and hyper-parameters.

The second column visualizes decomposition results.
On top, we report reconstruction errors (e.g., nre and rmse), indicating how well the given model fits the tensor and components similarity measured by FMS, indicating how distinct components are from each other.
We display top-$K$ factor values in a heatmap format to show the overall trend and highlight the entities having higher values.
Below the heatmaps, co-clusters across modes are summarized to show the dominant hidden patterns captured by each component.

The third column explains the co-clusters obtained from the second module. 
The first and second rows describe the latent patterns using task-agnostic and task-specific prompts.
We explain the results using metadata (author name, conference name, keywords, research area). 
For task-specific explanations, LLM is asked to predict future research trend topics based on co-clusters and their keywords.
The last row shows the results of forward and backward inference tasks of LLM generations. This is because forward inference requires the model to retrieve plausible entities from a larger candidate pool, whereas backward inference only involves detecting an inconsistent entity within a small set.

\vspace{-4mm}
\section{Conclusion}		\label{sec:conclusion}
\vspace{-3mm}
We introduce \method{} to facilitate the analysis of real-world tensors using LLMs.
Our system integrates interpretable tensor models, including CP and NeAT, with LLM-based interpretation to generate natural language explanations of latent patterns.
With task-agnostic and task-specific prompts, our system provides not only a general overview but also actionable insights.
Furthermore, forward and backward inference tasks improve the explanation credibility and quality of LLM-explanations.

\vspace{-5mm}
\section*{Acknowledgements}
\vspace{-2mm}
Research was supported in part by the National Science Foundation under CAREER grant no. IIS 2046086, by a UCR Senate Committee on Research grant, and by the Agriculture and Food Research Initiative Competitive Grant no. 2020-69012-31914 from the USDA National Institute of Food and Agriculture.
\vspace{-3mm}
\bibliographystyle{splncs04}
\vspace{-2mm}
\bibliography{dawon}
%
%
%
%

\end{document}